\documentclass[11pt,a4paper]{article}
\usepackage[hyperref]{acl2021}
\usepackage{times}
\usepackage{latexsym}

\usepackage{microtype}
\usepackage{booktabs}
\usepackage{multirow}
\usepackage{array,graphicx}
\usepackage{multicol}
\usepackage{float}
\restylefloat{table}
\usepackage{placeins}
\aclfinalcopy % Uncomment this line for the final submission

\newcommand*\rot{\rotatebox{90}}

\title{Do Language Models Perform Generalizable Commonsense Inference?}

\author{Peifeng Wang$^{1,2}$,\quad Filip Ilievski$^{2}$,\quad Muhao Chen$^{1,2}$,\quad Xiang Ren$^{1,2}$\\
$^{1}$Department of Computer Science, University of Southern California\\$^{2}$Information Sciences Institute, University of Southern California
\\
\texttt{\{peifengw,muhaoche,xiangren\}@usc.edu},\quad\texttt{ilievski@isi.edu}}

\begin{document}
\maketitle
\begin{abstract}
	Inspired by evidence that pretrained language models (LMs) encode commonsense knowledge,
recent work has applied LMs to automatically populate commonsense knowledge graphs (CKGs).
However, there is a lack of understanding on their generalization to multiple CKGs, unseen relations, and novel entities. This paper analyzes the ability of LMs to perform generalizable commonsense inference, in terms of \emph{knowledge capacity}, \emph{transferability}, and \emph{induction}.
Our experiments with these three aspects show that: (1) LMs can adapt to different schemas defined by multiple CKGs but fail to reuse the knowledge to generalize to new relations. (2) Adapted LMs generalize well to unseen subjects, but less so on novel objects. Future work should investigate how to improve the transferability and induction of commonsense mining from LMs.\footnote{\small The code is avaiable at \url{https://github.com/wangpf3/LM-for-CommonsenseInference}.} 
\end{abstract}

\section{Introduction}
\vspace{-0.1cm}
%Pretrained language models (LMs) have shown impressive performance on many NLP tasks, largely due to the tremendous knowledge captured from free text through pretraining (citations). 
%At the same time, l
Large-scale commonsense knowledge graphs (CKGs), like ConceptNet~\cite{speer2017conceptnet} and ATOMIC~\cite{sap2019atomic}, store structured knowledge that can benefit various knowledge-driven applications. Given the usefulness of CKGs, but also their inability to flexibly provide information,
%the expensiveness of constructing them
~\cite{paulheim2018much},
recent work has paid much attention to %shown promising results of %complementing
populating CKGs with commonsense knowledge mined from pretrained language models (LMs)~\cite{wang2020connecting,bosselut2019comet}. Enhancing the knowledge of CKGs is essential to support reasoning on downstream tasks~\cite{talmor2019commonsenseqa,wang2020joint,young2018augmenting}.

The task of completing CKGs has typically been posed as \textit{commonsense knowledge inference}, where the goal is to predict the {object} of a fact triplet, given its \emph{subject} and a \emph{relation} (\emph{predicate})~\cite{petroni2019language,bosselut2019comet}.
Commonsense inference techniques, such as COMET~\cite{bosselut2019comet}, typically fine-tune an LM, like GPT~\cite{radford2018improving}, over the training set from a single CKG. While such methods are able to dynamically enhance the completeness of CKGs, their application so far has been limited to the relation set of the source (training) CKG~\cite{da2021understanding}. In addition, the generated object concepts are found to be largely biased towards the ones in the training set~\cite{wang2020commonsense}. It remains unclear to which extent LMs can generalize to multiple CKGs, new relations, and novel objects.
To this end, we pose %conduct large-scale, empirical analysis to investigate
the question: \emph{do language models perform generalizable commonsense inference?}

\begin{figure}[t]
    \centering
    \includegraphics[width=0.49\textwidth]{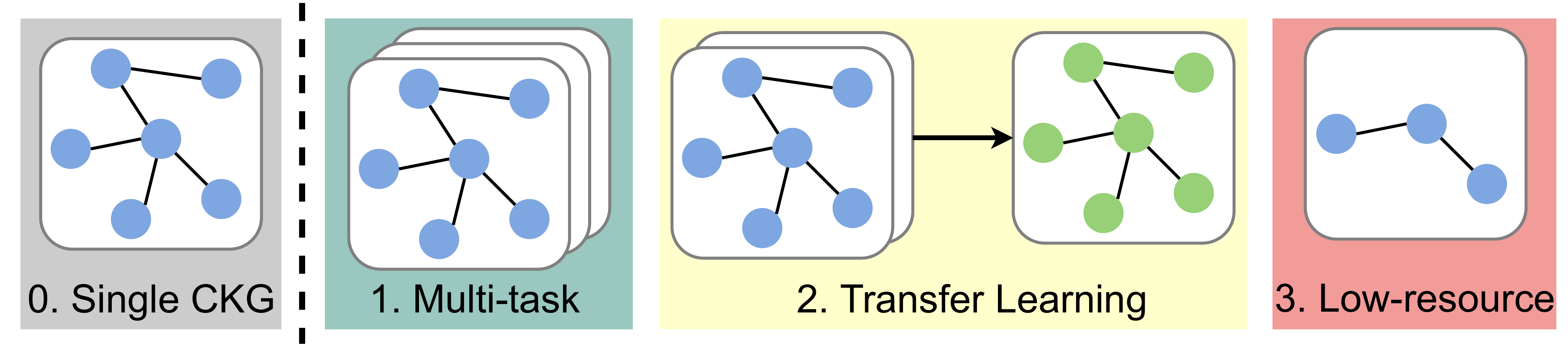}
\vspace{-0.8cm}
    \caption{Unlike previous studies that adapt LM on one single CKG (0), we investigate LM's three aspects of \textbf{generlizability}: (1) \textit{knowledge capacity} by multi-task learning, (2) \textit{transferability} by transfer learning and (3) \textit{induction} by controlled low-resource learning.}
    \label{fig:overview}
    \vspace{-0.3cm}
\end{figure}

To answer this question, we study three aspects of the LM generalizability for commonsense inference, namely: knowledge capacity, transferability, and induction. To measure the \textit{knowledge capacity} ability of LMs, we examine whether LMs can be adapted to multiple CKGs simultaneously, and tested on each of the CKGs. We test their \textit{transferability} by assessing whether an initial adaptation of a LM on multiple source CKGs can reduce the effort on further adapting it to a new CKG. The \textit{inductive} power of LMs is measured by varying the overlap between the objects in the training and test splits of a CKG. The overview of our analysis is depicted in Figure~\ref{fig:overview}. Our results show that LMs are able to infer knowledge for multiple CKGs simultaneously without loss of performance on the target inference task, though the transferability of knowledge across tasks is limited. In addition, we observe that the inductive power of LMs for commonsense inference relies heavily on whether an object is observed during training.

\section{Analysis Setup}\label{sec:background}
\vspace{-0.1cm}
% \xiang{Should start with re-stating the goal of the study, and overviewing the analysis questions. Then transit to the next subsection.}

To shed light on the LM's generalizalibility for commonsense inference, we %perform a series of systematic studies aiming to
investigate: whether LMs have the capability to adapt to multiple CKGs (\emph{Q1: capacity}), whether LMs can reuse the knowledge learned from source CKGs to efficiently adapt to a target CKG (\emph{Q2: transferability}), and %whether LMs are learning the relation schemas or are just memorizing the observed knowledge during adaptation 
whether LMs can predict unseen objects or mainly repeat the observed ones (\emph{Q3: induction}). In this Section, we define the task, the CKGs we consider, our experimental settings, and relate to prior studies.
% To answer these questions, we start by introducing the problem definition as follows. 

%Assume we are given a pre-trained LM together with an incomplete CKG. Our main goal is leveraging the commonsense triplets in the CKG as training examples to adapt %the neural representation of the LM towards commonsense inference. 
%Afterwards, we can adopt the adapted The goal of such adaptation is to leverage the improved LM %as a complementary to CKG for providing commonsense knowledge. to generate complementary knowledge to the CKG.

% \xiang{suggested structure:

% 0. problem definition?

% 1. what are the couple analysis questions and the corresponding exp settings?

% 2. what datasets/KGs we look at?

% 3. evaluation protocols/metrics?

% Right now 1/2 are mixed in Sec 2.1 and even in Sec 4.2/4.3, and 3 is sort of unclear/missing}

\subsection{Task Formulation}
Following%prior studies
~\citet{hwang2020comet,da2021understanding}, we formalize \textit{commonsense inference} 
as a task of predicting the object of a triplet, given a pair of \textit{(subject, relation)} as input. The subject $s$ and the object $o$ are both expressed as free-form phrases, while the relation $r$ is a predefined relation type from the CKG. A training example from ConceptNet could have \texttt{(go to a concert, MotivatedByGoal)} as input, and \texttt{listen to music} as output. Assuming that a CKG is given, the goal is to leverage the commonsense triplets in the CKG as training examples to adapt the LM for commonsense inference.

\subsection{CKG Datasets}
%We consider the CKGs among different fields of commonsense knowledge as follows.

%As datasets, 
We consider three large and popular CKGs, with different foci:% following CKGs that differ in domains. 
(1) \textbf{ConceptNet}'s broad set of commonsense knowledge includes taxonomic (e.g., \texttt{IsA}), utility (e.g., \texttt{UsedFor}), and temporal knowledge (e.g., \texttt{HasPrerequisite}). It combines %constructed by both 
crowdsourced knowledge with that from existing sources, such as WordNet. % and Wiktionary. % and expert editors. 
%In particular, w
We use its ConceptNet-100K subset, collected by~\citet{li2016commonsense}. (2) \textbf{TupleKB}~\cite{Mishra2017DomainTargetedHP} focuses on scientific commonsense knowledge like \texttt{(salt, dissolve in, water)}. It is constructed through an information extraction pipeline. (3) \textbf{ATOMIC}~\cite{sap2019atomic} has social commonsense knowledge about causes and effects of everyday events, and mental states (e.g., \texttt{xIntent}) of their participants. It is created by crowdsourcing.

As indicated by~\citet{jastrzkebski2018commonsense}, a large proportion of the subjects in the test set of ConceptNet-100K overlap with its training set, while TupleKB does not provide an official split.
%Considering these issues
Thus, we (re-)split these two datasets to ensure  %following the criterion 
that the subjects of testing triplets do not appear in the training set. This criterion is also consistent with how the ATOMIC dataset is constructed.

\subsection{Experimental Settings}
%The overview of our analysis is depicted in Figure~\ref{fig:overview} and the experiments are introduced as follows.
% \vspace{-0.1cm}

\smallskip
\noindent
\textbf{Multi-task Learning} To answer Q1, we adapt an LM with balanced training data from ConceptNet, TupleKB, and ATOMIC. %To balance between each
%CKG, w
We sample 8 triplets from each dataset to form one training batch. 
%As introduced above, each input triplet is converted into natural language phrases, and the conversion obfuscates the information about the source CKG of the triplet. % which CKG the triplet belongs to. 
%We include two baselines in this experiment for comparison, both of which use GPT2-XL in a zero-shot setting. The difference is that one is fed with the same input as our Fine-tuning method while the other is fed with the input with demonstration as our FT+demonstration method.

\smallskip
\noindent
\textbf{Transfer Learning}~To provide insight into Q2, we adopt 
%two learning settings. One is still multi-task learning as our first study, where the LM gets access to all the CKGs. The second one is 
transfer learning under a leave-one-out strategy. In this setting, we adapt an LM on two of the three CKGs, and then we further adapt it on the third target CKG. Moreover, we %look into low-resource learning scenarios 
study the data efficiency of this transfer learning by down-sampling each training set to $x=\{1,20,50\}\%$,
in order to see whether the LM can adapt to the target CKG with less training effort. Fine-tuning on data as small as 1\% training set may suffer from instability, and results may change dramatically given a new split of training data~\cite{gao2020making}. To control the randomness, we re-sample the 1\% training data 5 times with a fixed set of random seeds and report the average performance instead. 

\smallskip
\noindent
\textbf{Controlled Low-resource Learning}~To answer Q3, we design a controlled experiment, where we first split the training set into two disjoint subsets %distinguishing
depending on whether the triplets in the original training set contain objects that exist in the test set or not. We denote the subset where the objects of the triplets appear in testing data as $\Omega$. We sample $x=\{0,25,50,100\}\%$ of the training triplets in $\Omega$ for adapting the LM. During the evaluation, we also separate the test set into two disjoint subsets, according to whether the objects are seen in the original full training set. The results on these two split test sets are reported separately for each adapted LM.

\smallskip
\noindent
\textbf{Evaluation Protocol}
For each \textit{(subject, relation)} pair in the test set, we treat all their objects as ground truth references for evaluating the model inference. %by the models.
We report scores for commonly used automatic evaluation metrics for text generation: BLEU~\cite{papineni-etal-2002-bleu}, ROUGE~\cite{lin2004rouge}, and METEOR~\cite{banerjee2005meteor}, which are shown to be consistent with human judgements~\cite{hwang2020comet}. During experiments, we observe a high correlation among these different metrics and choose to report METEOR in the main text and other metrics in the appendix.

\subsection{Connections to Prior Studies}

Earlier works~\cite{li2016commonsense,jastrzkebski2018commonsense,davison2019commonsense} poses the CKG completion task as triplet classification, where the goal is to score the plausibility of a complete triplet. COMET~\cite{bosselut2019comet} is the first to cast this task as commonsense inference with LMs. Follow-up contributions %extend~\citeauthor{bosselut2019comet}'s study by 
utilize COMET as a commonsense provider in various downstream tasks~\cite{bosselut2019dynamic,ammanabrolu2020automated,chakrabarty2020generating}, thus providing evidence for LM's generalization to previously unseen scenarios. Further efforts include~\citet{hwang2020comet}, which show that the quality of the training triplets is a key factor of adapting LMs, and~\cite{da2021understanding}, which investigates how to learn COMET in a few-shot learning setting. Meanwhile, the study by~\citet{wang2020commonsense} indicates the limited generalization of COMET.
\citet{Ma2021} also adapt LMs simultaneously on multiple CKGs, albeit their goal is to improve downstream performance rather than CKG inference. In this paper, we aim to provide a more comprehensive study of a LM's generalizability for CKG inference.%, in this paper we evaluate three aspects of this challenge: knowledge capacity, transferability, and induction. 

\section{Method}
%\xiang{can we make a figure to illustrate/compare all methods?}

\begin{table}[t]
\center
\scalebox{0.7}{
\begin{tabular}{l|cc}
\toprule
\textbf{Adaptation method} & \textbf{Input} & \textbf{Learnable params} \\\midrule
Zero-shot (ZS) & $(s, r)$ & N/A \\
ZS+demo & $(s^{'},r,o^{'},s,r)$ & N/A \\
Fine-tuning (FT) & $(s, r)$ & Transformer (LM) \\
FT+demo& $(s^{'},r,o^{'},s,r)$&Transformer (LM)\\
Adapter tuning (AT) & $(s, r)$ & Adapter \\\bottomrule
\end{tabular}
}
\vspace{-0.1cm}
\caption{
%Comparison of different m
Methods for using LMs to conduct commonsense inference. %ZS is short for ``zero-shot'' and FT is short for ``fine-tuning''. 
\emph{``+demo''} means prepending a demonstration triplet $(s^{'},r,o^{'})$ before the input tuple.\vspace{-0.2cm}
}
\label{tab:method}
\end{table}

While a set of pretrained LMs exists,
we adopt a widely used generative model, GPT2~\cite{radford2019language}, as our baseline LM.
%we focus on a generative LM which meets our need for generating knowledge. In particular, we adopt the GPT2 model~\cite{radford2019language} as our baseline LM, while 
The investigation of other generative LMs is orthogonal to our analysis. We experiment with its largest version, GPT2-XL, which contains 48 transformer layers~\cite{vaswani2017attention}, ensuring sufficient capacity for storing knowledge acquired during its pretraining. We introduce our experimental method as follows.

% \subsection{Commonsense as Natural Language}
\smallskip
\noindent
\textbf{Commonsense Inference with LMs}
%One advantage of leveraging a LM for conducting commonsense inference is that it provides a more accessible interface for any query expressed in natural language. 
Given a training triplet \textit{(s,r,o)}, 
we represent $s$ and $o$ as sequences of tokens, $\textbf{x}_s$ and $\textbf{x}_o$, which is trivial given that they are already expressed as phrases. As for the relation $r$, we convert it by using a template taken from the literature~\cite{davison2019commonsense} into a natural-language phrase $\textbf{x}_r$, e.g., \texttt{IsA} is converted to ``is a''. This has been shown to facilitate efficient adaptation of LMs~\cite{da2021understanding}. Note that we do not explicitly provide the LMs with the information about the source CKG of the triplet as input (e.g., prepending a related special token to the triplet).

%We do not focus on designing better templates since it has been investigated in the previous work~\cite{jiang2020can}.

\smallskip
\noindent
\textbf{Adapting LMs with Commonense Knowledge}
The training objectives for adapting LMs is to maximize the probability of generating the object phrase $\textbf{x}_o$ given the tuple $(\textbf{x}_s, \textbf{x}_r)$. %The loss function can formally be defined as: ${\mathcal{L}=-\sum_{t=|\textbf{x}_s|+|\textbf{x}_r|}^{|\textbf{x}_s|+|\textbf{x}_r|+|\textbf{x}_o|}\log P(x_t|x_{<t})}$.
During inference, we adopt greedy decoding to obtain the predicted object from the adapted LM.

%\subsection{Adaptation Paradigms}
There have been various techniques developed for adapting pretrained LMs to downstream tasks~\cite{howard2018universal,chen2020recall}. Moreover, previously only the vanilla \textbf{Fine-tuning}, i.e., updating the whole LM architecture during training, has been employed to adapt LMs for commonsense inference~\cite{bosselut2019comet,hwang2020comet,da2021understanding}. To obtain comprehensive results that are not specific to one particular way of fine-tuning, here we investigate two more alternatives, each of which has their own advantage when considered in different contexts.

\begin{figure*}[!ht]
\vspace{-0.1cm}
    \centering
    \includegraphics[width=0.92\textwidth]{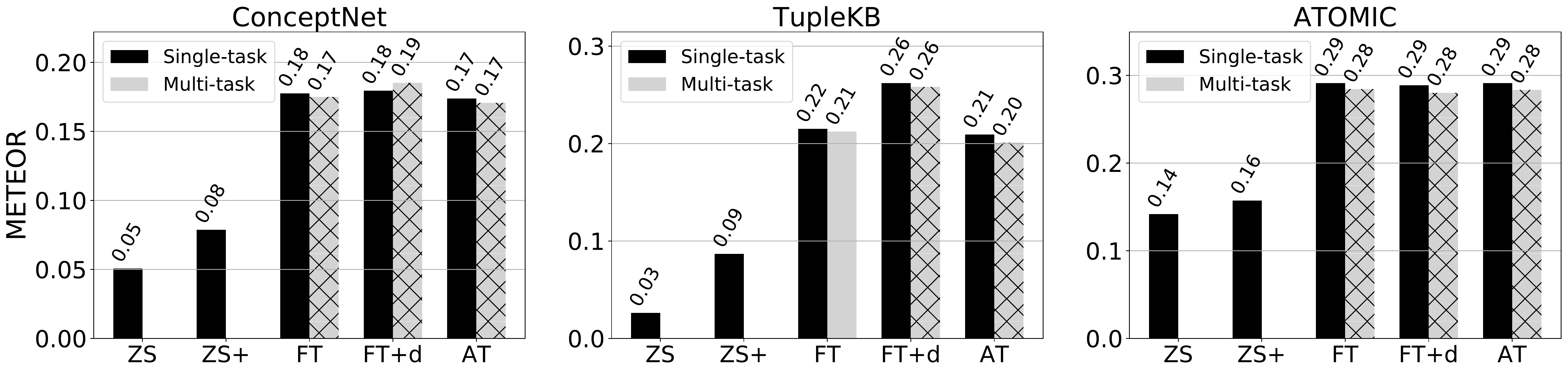}
    \vspace{-0.2cm}
    \caption{\small \textbf{Results (METEOR) for \textit{knowledge capacity} of LMs. "FT+d" refers to FT+demo.} We find no notable performance drop for any method trained in the multi-task setting.}
    \label{fig:multi-task}
    \vspace{-0.1cm}
\end{figure*}

%Existing studies~\cite{bosselut2019comet,hwang2020comet,da2021understanding} on adapting LMs for commonsense inference learns one model for each CKG. This assumes that the optimal CKG for a downstream task is known in advance, which is typically not the case. %leads to a requirement for the model developer to know in advance what type of knowledge is needed by the downstream task. For this reason, we examine the LMs' capability to adapt to multiple CKGs in varied domains at once. Besides being more realistic, this setup is also more parameter-efficient.

\begin{figure}[!ht]
    \hspace{-0.2cm}
    \includegraphics[width=0.49\textwidth]{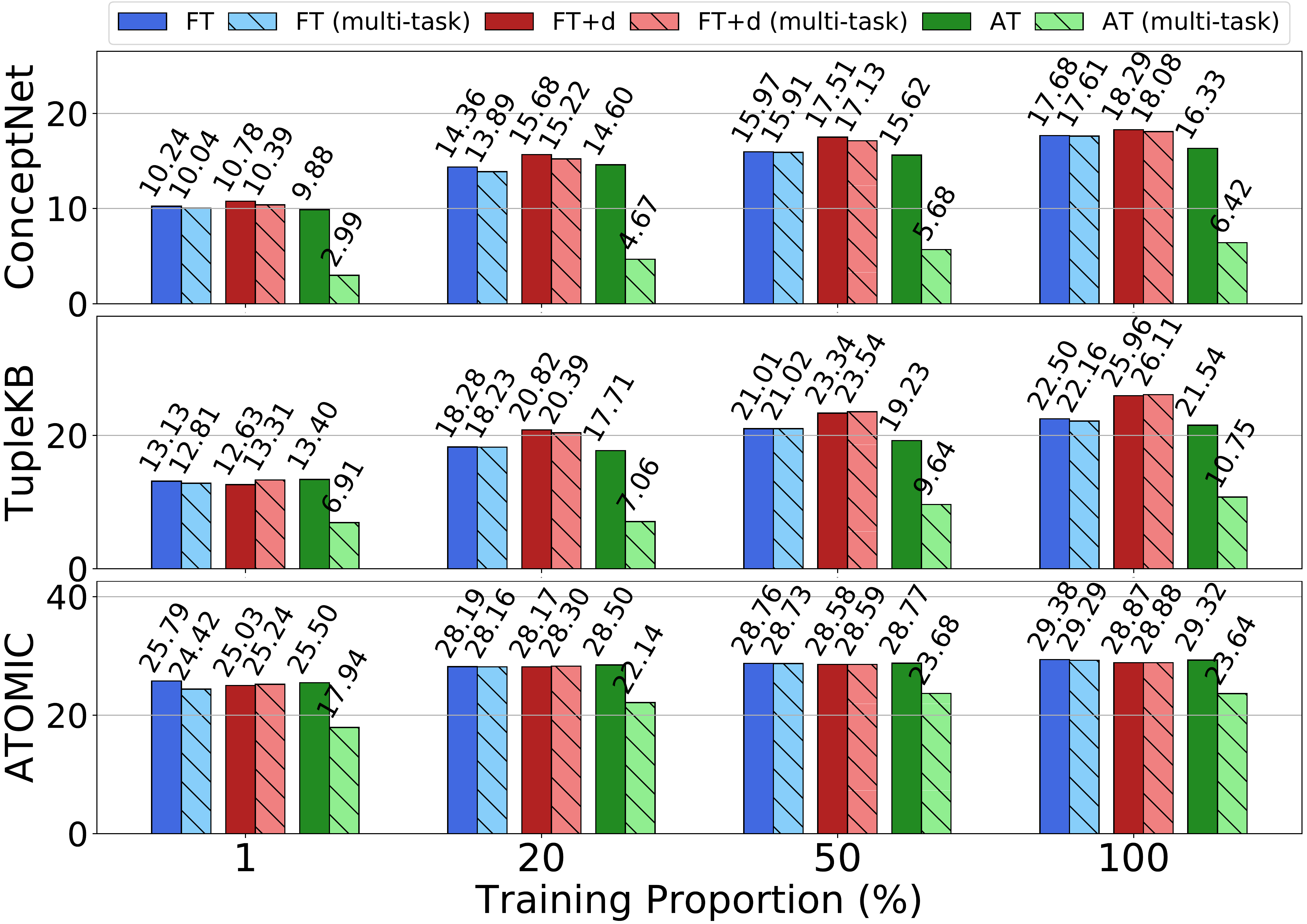}
    \vspace{-0.6cm}
    \caption{\small \textbf{Results (METEOR) for LM \textit{transferability}.} "FT+d" refers to FT+demo. Across datasets, we do not observe that adapting to the source CKGs would enable the LMs to adapt to the target CKG better or more easily. %, with blue lines for the single-task results and red lines for the transfer-learning results.
    }
    \label{fig:transfer}
\end{figure}

\begin{figure}[!ht]
    \hspace{-0.2cm}
    \includegraphics[width=0.49\textwidth]{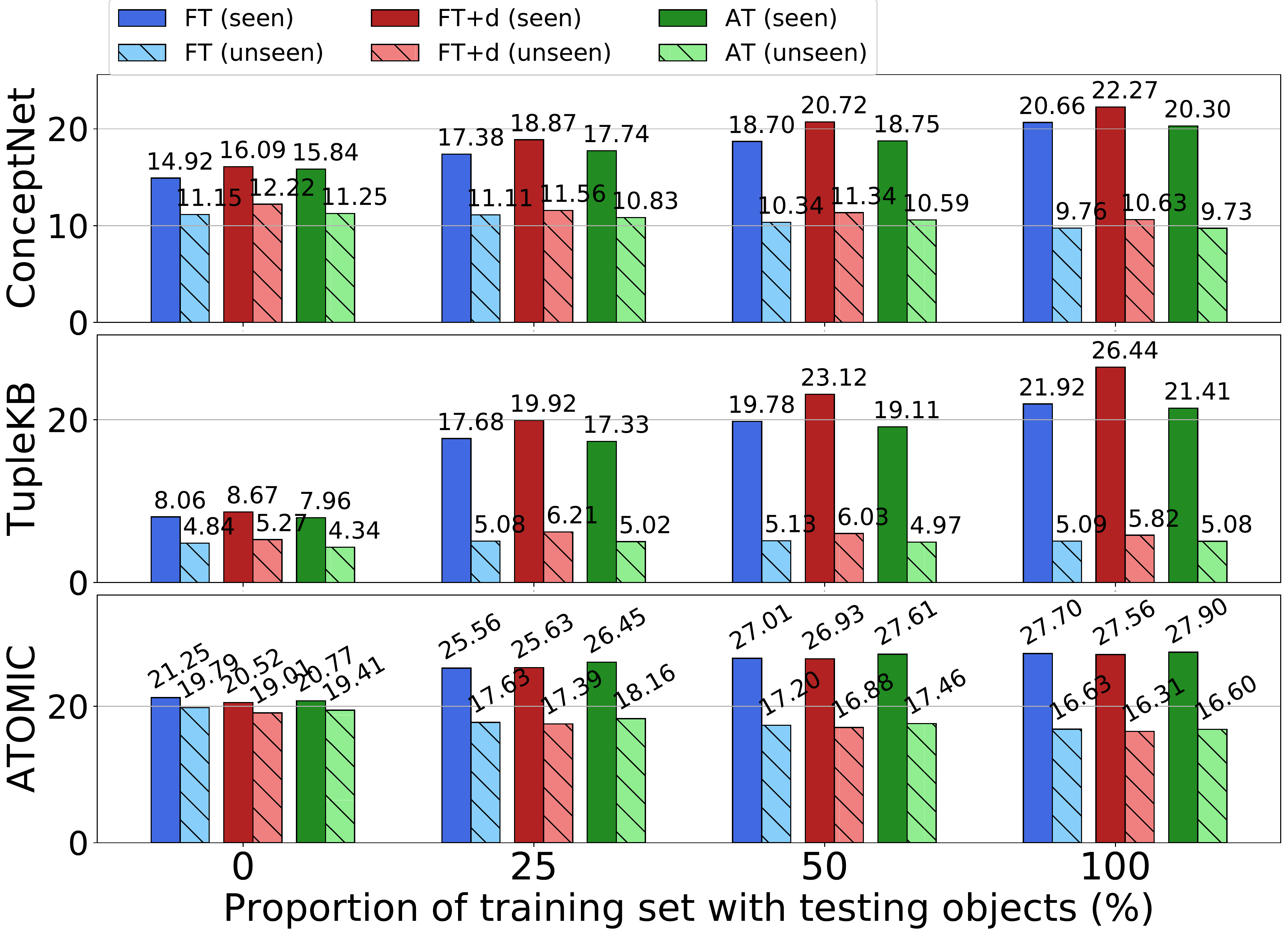}
    \vspace{-0.6cm}
    \caption{\small \textbf{Results (METEOR) for LM \textit{induction}.}  "FT+d" refers to FT+demo. All the methods perform better on predicting facts that contain seen objects, while the performance degrades when less objects are seen during training. %The blue lines denote the test subset that contains training objects, while the red ones denote the test subset that does not contain training objects.
    }
    \label{fig:low_resource}
\vspace{-0.3cm}
\end{figure}

\smallskip
\noindent
\textbf{Fine-tuning with Demonstration (FT+demo)} Combining the ideas of fine-tuning and in-context learning~\cite{brown2020language}, this technique~\cite{gao2020making} adds a demonstration to each input as additional context and fine-tunes the whole LM as usual. Incorporating demonstrations is shown to boost performance when the amount of training data is extremely limited. In our case, a demonstration is a top-1 training triplet $(s^{'},r,o^{'})$, ranked according to the cosine similarity between the embedding of the input tuple $(s,r)$ and the embeddings of the training tuples with the same relation type $r$. The tuple embeddings are given by a pretrained Sentence-BERT~\cite{reimers2019sentence}. For instance, a demonstration (\texttt{go to restaurant}, \texttt{UsedFor}, \texttt{eat out}) would be added before the input (\texttt{go to pub}, \texttt{UsedFor}). %as a complete input to the LM.
With the demonstrated triplets, the LM could learn to understand the schema of the CKG instead of simply learning the knowledge from the training data. 

\smallskip
\noindent
\textbf{Adapter Tuning (AT)}~Unlike fine-tuning, adapter tuning~\cite{houlsby2019parameter} fixes the entire LM and adds one trainable adapter right before the skip connection in each transformer layer of the LM, which is more parameter-efficient. Each adapter is a two-layer bottleneck network with a skip-connection internally. Following \citet{houlsby2019parameter}, the parameters of the bottleneck network are initialized close to zero so that the adapter approximates an identity function from the beginning. %We refer readers to~\citet{houlsby2019parameter} for more details.

We compare to two additional baselines, both using GPT2-XL in a zero-shot setting: \textbf{Zero-shot (ZS)} is fed with the same input as Fine-tuning, while zero-shot with demonstrations (\textbf{ZS+demo}) combines the input plus demonstration, as in the FT+demo method. By investigating all these methods, we aim to understand the influence of different adaptation techniques on the models' performance. Table~\ref{tab:method} summarizes the set of methods which we consider in this paper.

\section{Results and Discussion}

% \smallskip
\noindent
\textbf{Knowledge Capacity (Q1)}
The results that quantify the knowledge capacity of LMs for commonsense inference over multiple CKGs with METEOR scores are shown in Figure~\ref{fig:multi-task}. The complete results including other metrics can be found in the appendix.  All adaptation methods perform considerably better than the zero-shot baselines, indicating the benefit of adaptation. There is no clear distinction between the adaptation methods, though FT+demo performs slightly better than the others across CKGs. Most importantly, we find no notable performance drop for any method in the multi-task training setup despite the challenge that there is limited overlap between these CKGs. Only $10.0\%$ of the facts from ATOMIC can be found in ConceptNet~\cite{hwang2020comet} while $8.4\%$ of the facts from ConceptNet can be found in TupleKB~\cite{Mishra2017DomainTargetedHP}~\footnote{We also try to breakdown the results by relation types and do not observe correlation between the relation-wise performance and the extent of overlap.}. % that each method does not lose much performance when they are trained under multi-task, 
This indicates the prominent capacity of LMs to simultaneously adapt to different CKGs. Nevertheless, the results reveal that learning different CKGs jointly do not interfere with each other positively (via knowledge sharing) or negatively (due to overfitting).

\smallskip
\noindent
\textbf{Transferability (Q2)}
%Given that LMs can generalize to multiple CKGs, we further study whether the learning on some CKGs can bring improvement to the inference on others.
Figure~\ref{fig:transfer} shows the obtained results regarding the transferability of LMs. Across different CKGs and for any training data size, we observe no indications that % turn out to be negative. Among different CKGs, we do not find the
adapting to the source CKGs enhances  the performance on the target CKG. %even with less training data. 
On the contrary, adapting from source CKGs even hurts the performance of the Adapter-tuning method, revealing that this method overfits to the source CKGs. Overall, we conclude that LMs cannot reuse the knowledge learned from the source CKGs to improve the performance on the target CKG or achieve the same performance with less training data. Thus, we call for future study on developing more effective adaptation methods.
%these CKGs are largely disjoint and the knowledge from one source cannot facilitate the adaptation to another source.

\smallskip
\noindent
\textbf{Induction (Q3)}
%In our last study, we examine what exactly is picked up by LMs during adaption. In our first two studies, we observe LMs' capability to accommodate diverse commonsense schema defined by multiple CKGs, while failing to transfer the knowledge across CKGs. Therefore, we hypothesize that LMs are only adapted to the knowledge that they observe during training.
The results in Figure~\ref{fig:low_resource} show that without down-sampling ($x=100\%$), all methods perform much better on predicting facts that contain seen objects, and their performance %gradually drop when less objects have been observed
degrades more when less object entities are seen to training. Meanwhile, the performance on %test subset
facts %that do not contain observed
with unseen objects stays roughly unaffected. This indicates a key limitation of the LMs: they adapt notably better on seen objects. Since the training set and test set do not share subjects, we conclude that the generalizability of the LM is largely dependent on
finding the relationship between unseen subjects and observed objects. We thus posit that a novel strategy for adapting LMs while retaining the knowledge acquired during pre-training is necessary for better generalizability. Promising directions here are prefix tuning~\cite{li2021prefix} or including an additional objective during adaptation which would encourage the generation of novel objects.

\section{Conclusion}
\vspace{-0.2cm}
This work conducted a focused study of
%a systematic study in order to understand 
three aspects of the generalizability of LMs for commonsense inference: knowledge capacity, transferability, and induction. We experiment with five methods of using a generative LM and three representative CKGs. Despite their capability to accommodate multiple CKGs, we have observed that LMs %its limitation in 
have limited ability to transfer knowledge across CKGs. Moreover, their adaptation relies heavily on whether the objects to predict are seen during training. These findings help our understanding of LMs' adaptation behavior on commonsense inference, and highlight the need for future work to improve their %limited
transferability and induction.
\section*{Acknowledgments}
We thank the anonymous reviewers for their insightful comments. This material is based upon work sponsored by the DARPA MCS program under Contract No. N660011924033 with the United States Office Of Naval Research.

\bibliography{acl2021}
\bibliographystyle{acl_natbib}

\appendix
% !TEX root = main.tex
% \begin{multicols}{2}
% [\section{Additional Results}]
% \end{multicols}

\begin{table*}[hbt!]
\small
\center
\begin{tabular}{llrrrrrr}
\toprule
\multirow{2}{*}{}&\multirow{2}{*}{} & \multicolumn{2}{c}{BLEU-2} & \multicolumn{2}{c}{ROUGE-L} & \multicolumn{2}{c}{METEOR} \\ \cmidrule(l){3-8} 
& & single-task & multi-task & single-task & multi-task & single-task & multi-task \\ \midrule
&Zero-shot & 0.0069 & NA & 0.1009 &NA  & 0.0506 & NA \\ 
&ZS+demo & 0.0284 &NA  & 0.1281 & NA & 0.0787 &  NA\\
&Adapter-tuning & 0.1289 & 0.1279 & 0.2598 & 0.2560 & 0.1739 & 0.1706 \\ 
&Fine-tuning & 0.1325 &0.1286  & 0.2629 & 0.2575 & 0.1775 &0.1749  \\ 
\rot{\rlap{~ConceptNet}}&FT+demo &\textbf{0.1333}  & \textbf{0.1398} & \textbf{0.2678} & \textbf{0.2738} & \textbf{0.1795} & \textbf{0.1851} \\ \midrule
&Zero-shot & 0.0017 &NA  &0.0999  & NA & 0.0263 &NA  \\ 
&ZS+demo & 0.0099 & NA &0.2748 &  NA&  0.0869& NA \\
&Adapter-tuning & 0.1383 &0.1323  & 0.3785 & 0.3627 &0.2094  & 0.2010 \\ 
&Fine-tuning & 0.1371 & 0.1388 & 0.3985 & 0.3812 & 0.2151 & 0.2122 \\ 
\rot{\rlap{~TupleKB}}&FT+demo & \textbf{0.1699} & \textbf{0.1698} &  \textbf{0.4902}& \textbf{0.4714} & \textbf{0.2622}& \textbf{0.2580} \\\midrule &Zero-shot & 0.0436 & NA & 0.2523 & NA & 0.1419 & NA \\ 
&ZS+demo & 0.0808 & NA & 0.2233 & NA & 0.1572 & NA \\
&Adapter-tuning & \textbf{0.2161} & 0.2035 & \textbf{0.4008} &0.3890  & \textbf{0.2913} & 0.2832 \\ 
&Fine-tuning & 0.2125 & 0.2057 & 0.3982 & \textbf{0.3908} & 0.2913 & \textbf{0.2843} \\ 
\rot{\rlap{~ATOMIC}}&FT+demo & 0.2111 & \textbf{0.2070} & 0.3915 & 0.3868 & 0.2887 & 0.2800 \\ \bottomrule
\end{tabular}
\caption{Results of all the evaluation metrics for the knowledge capacity experiments.}
\label{tab:multi-task}
\end{table*}
\FloatBarrier
\section{Appendix}
\subsection{Dataset Statistics}

\begin{table}[h]
\centering
\small
\caption{CKG Dataset Statistics.}
\label{tab:dataset_stat}[h]
\begin{tabular}{lrrr}
\toprule        & Train & Dev   & Test  \\
\midrule
ConceptNet100k&79,770&10,203&10,027\\
TupleKB&98,674&12,357&12,427 \\
ATOMIC &578,002&64,902&71,127\\
\bottomrule
\end{tabular}
\end{table}
\subsection{Implementation Details}
The GPT2-XL language model we adopted in this work has 1558M parameters in total. We train all the models on a V100 GPU. As for hyper-parameters, we adopt the commonly-used learning rate (1e-5) and batch size (16) for adapting GPT2, except that in the multi-task learning setting, the batch size is 24 (8 samples from each CKG).

\subsection{Additional Results}
See Table~\ref{tab:multi-task} for the full results of all the evaluation metrics considered in this paper.

\end{document}